\documentclass[10pt,twocolumn,letterpaper]{article}

\usepackage[pagenumbers]{wacv}
\usepackage[T1]{fontenc}
\usepackage{bm}
\usepackage{pifont}
\usepackage{bbding}

\newcommand{\cmark}{\ding{51}}
\newcommand{\xmark}{\ding{55}}
\newcommand{\vstd}[2]{#1{\scriptsize$\pm$#2}}

\definecolor{wacvblue}{rgb}{0.21,0.49,0.74}
\usepackage[breaklinks,colorlinks,allcolors=wacvblue]{hyperref}

\def\wacvPaperID{3381}
\def\confName{WACV}
\def\confYear{2027}

\title{MoWE: Spatially Adaptive Mixture-of-Wavelet-Experts Pooling for Ultrasound Lesion Segmentation}
\author{Reaz Hassan Joarder$^{1}$ \quad Sidratul Tanzila Tasmi$^{2}$ \quad Alif Arshad Bakshi$^{1}$\\
Md.~Bakhtiar Hasan$^{1}$ \quad Md Hasanul Kabir$^{1}$\\[4pt]
$^{1}$Islamic University of Technology, Bangladesh \quad $^{2}$United International University, Bangladesh
}

\begin{document}
\maketitle
\begin{abstract}
  We present Mixture-of-Wavelet-Experts (MoWE), a spatially adaptive routing mechanism that replaces max-pooling and strided convolutions in U-Net architectures for ultrasound lesion segmentation. Standard pooling operations suppress high-frequency spatial details, including potential boundary cues in speckled regions. Recent frequency-domain methods preserve some of these details, but a single wavelet family applies the same spectral bias across the image. MoWE instead uses a context-aware router to assign local regions to fixed wavelet families according to local texture and global feature context. The decoder reuses the encoder's spatial routing maps during the inverse transform. Learnable channel-wise gates rescale the high-frequency subbands before reconstruction. We evaluate MoWE with U-Net on four ultrasound lesion benchmarks (BUSI, TN3K, UDIAT, BUS-UCLM) and one cardiac dataset (CAMUS), and with nnU-Net on BUSI. On BUSI, MoWE improves U-Net Dice from 72.59\% to 74.55\% and reduces HD95 from 47.01 to 43.49\,px. With nnU-Net, MoWE improves Dice from 80.18\% to 81.97\% and reduces ASSD from 20.07 to 17.25\,px. On U-Net, MoWE reduces trainable parameters by 6.4\%.

\end{abstract}

\section{Introduction}
Ultrasound (US) lesion segmentation supports non-invasive diagnosis by delineating pathological tissue from its surroundings. This task remains difficult because low tissue contrast, acoustic shadowing, imaging artifacts, and speckle can obscure lesion margins and create lesion-like patterns in adjacent tissue~\cite{huang2025ultrasound,aumente2024multitask}. A useful segmentation representation must therefore retain weak boundary evidence while avoiding indiscriminate amplification of high-frequency noise.

Resolution transitions are a consequential part of this problem. U-Net~\cite{ronneberger2015unet} and its self-configuring extension nnU-Net~\cite{isensee2021nnu} are strong reference architectures for medical image segmentation~\cite{isensee2024nnunet_revisited}, but their encoders repeatedly reduce feature resolution through pooling or strided convolutions. Such operations expand the effective receptive field and reduce computation, yet can remove spatial detail needed for dense prediction~\cite{el2025toward}. The transition is more than an implementation detail because it determines which information reaches later encoder stages. Discrete wavelet transforms provide an alternative by decomposing features into low-frequency and directional detail subbands~\cite{alessio2015discrete}. Recent segmentation networks exploit this decomposition for multi-scale representation, frequency-aware downsampling, or selective subband retention~\cite{qin2024mscwnet,gan2024wtcm,alhasan2025waveformer,ahmed2025wavegms,zhang2025uwtnet,el2025toward}. However, these designs choose the wavelet transform or retained subbands at the model or image level rather than selecting a wavelet family independently for each spatial region.

Our premise is that the appropriate frequency decomposition can vary within one US image. Lesion interiors, boundary segments, surrounding tissue, and shadowed regions exhibit different local structures, while wavelet families encode different trade-offs between spatial support and regularity~\cite{daubechies1992ten}. We therefore propose Mixture-of-Wavelet-Experts (MoWE), which routes each pooled spatial location among a small bank of fixed wavelet families. A convolutional router combines local texture with global feature context and produces discrete assignments through straight-through Gumbel-Softmax~\cite{jang2017gumbel}. The decoder reuses these assignments for inverse transformation and applies learned channel-wise gates to the cached detail subbands. This construction retains the known analysis--synthesis structure of each expert while making the basis choice spatially dependent.

We evaluate MoWE within U-Net and nnU-Net to test this routing-aware decomposition across established backbones. With U-Net, it improves Dice and HD95 on BUSI, BUS-UCLM, and TN3K; with nnU-Net on BUSI, it improves Dice from 80.18\% to 81.97\% and reduces ASSD from 20.07 to 17.25\,px.

\subsection{Related Work}
\textit{Encoder--decoder segmentation:} U-Net~\cite{ronneberger2015unet} established the encoder--decoder paradigm with skip connections for medical image segmentation. nnU-Net~\cite{isensee2021nnu} extended this architecture with a self-configuring training pipeline and provided strong results across the Medical Segmentation Decathlon~\cite{isensee2024nnunet_revisited,jiangtao2025comprehensive}. Attention gates~\cite{oktay2018attention}, transformer blocks~\cite{chen2021transunet,cao2022swinunet}, and state-space models~\cite{ma2024umamba} add alternative forms of feature refinement and context modeling. These encoder--decoder designs generally retain max-pooling or strided convolutions for downsampling, in contrast to the frequency-aware transitions reviewed below.

\textit{Downsampling and aliasing:} Max-pooling and strided convolutions subsample features without an explicit anti-aliasing filter, which breaks shift invariance~\cite{azulay2019why} and discards high-frequency content. Blur-pooling inserts a fixed low-pass filter before subsampling and improves stability under small transformations~\cite{zhang2019shift}, while wavelet pooling replaces learned pooling with a fixed subband decomposition~\cite{williams2018wavelet}. These methods apply one predetermined frequency response at every spatial location. MoWE instead selects among several fixed analysis--synthesis operators per location, so the effective frequency response adapts to local content while every candidate operator retains perfect reconstruction.

\textit{Wavelet-based segmentation:} Wavelet transforms address this by decomposing features into frequency subbands, preserving high-frequency details that pooling discards. Recent methods integrate wavelets into encoder--decoder architectures for multi-scale feature extraction~\cite{qin2024mscwnet,gan2024wtcm}, parameter-efficient downsampling~\cite{alhasan2025waveformer,ahmed2025wavegms}, and low-frequency feature mining~\cite{zhang2025uwtnet,liang2026wkpnet}. Matched Wavelet Pooling~\cite{el2025toward} selects task-optimal subbands but applies a single wavelet family globally. In all cases, the wavelet basis is fixed across the entire image, imposing a uniform spectral bias on heterogeneous US tissue, where homogeneous regions may favor smooth wavelets (e.g., db4) and sharp lesion boundaries may favor compactly supported ones (e.g., Haar).

\textit{Mixture-of-Experts in medical imaging:} Mixture-of-Experts (MoE) architectures route inputs to specialized sub-networks
via learned gating functions \cite{shazeer2017outrageously}. In medical imaging, MoE has been applied to route
encoder features through expert refinement layers~\cite{li2025moesam},
handle domain shift across 3D segmentation tasks~\cite{wang2024sam}, and
dynamically select scanning strategies in state-space
models~\cite{zhang2025switchumamba}. However, these methods route tokens or
feature channels to learned neural sub-networks. No prior work routes
\emph{spatial regions} to \emph{signal-processing operators}, specifically fixed wavelet
families with known frequency characteristics, making the gating decision
interpretable as a locally appropriate frequency decomposition.
\subsection{Our Contributions}
\begin{enumerate}
  \item We propose MoWE, a spatially adaptive routing mechanism that replaces standard pooling by dynamically assigning each local region to an appropriate fixed wavelet family, conditioned on both local texture and global anatomical context via Gumbel-Softmax gating.
  \item We introduce a detail-gated decoder that reuses encoder routing maps to guarantee analysis--synthesis consistency and employs learnable channel-wise gains to rescale detail subbands prior to reconstruction.
  \item We evaluate MoWE with U-Net across four US lesion benchmarks and one cardiac dataset, and with nnU-Net on BUSI.
\end{enumerate}

\begin{figure*}[t]
  \centering
  \includegraphics[width=\textwidth]{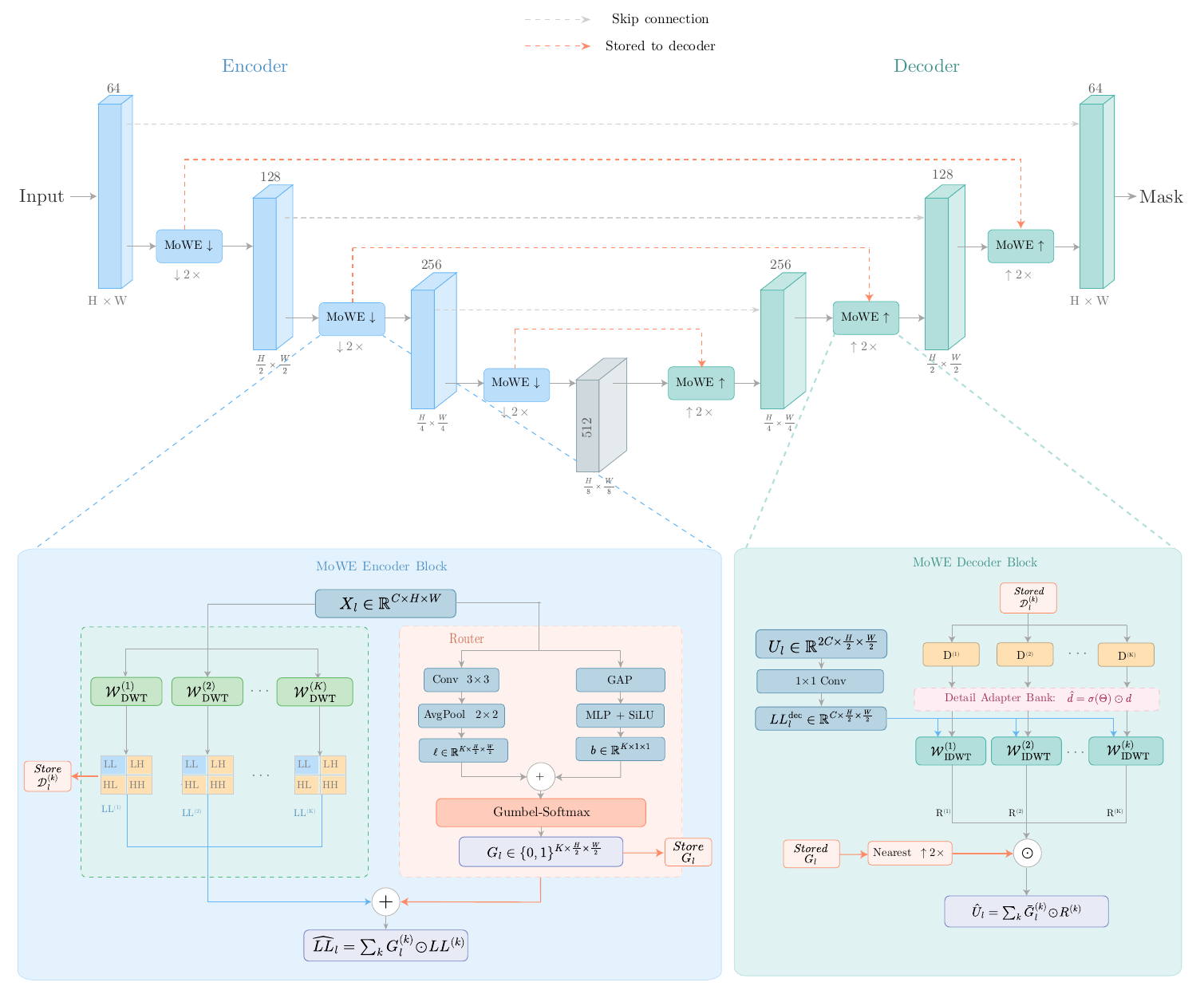}
  \caption{MoWE: Context-aware spatial wavelet routing and detail-gated reconstruction.}
  \label{fig:mowe_architecture}
\end{figure*}

\section{Methods}\label{sec:method}

MoWE (Fig.~\ref{fig:mowe_architecture}) consists of a context-aware encoder router, a decoder that reuses the routing maps with learned detail gating, and a load-balancing term for the expert assignments.

\subsection{Backbone Integration and Tensor Shapes}\label{sec:architecture}
MoWE replaces the resolution transitions of the encoder--decoder. The U-Net
configuration has four wavelet transitions with channel widths
$64\!\rightarrow\!128\!\rightarrow\!256\!\rightarrow\!512\!\rightarrow\!1024$. For nnU-Net,
the stage depths and channel widths follow its automatically selected 2D
configuration.
At level $l$, the feature tensor entering a transition has shape
$B\times C_l\times H_l\times W_l$. Every fixed expert applies a depthwise,
stride-two DWT and returns four tensors of shape
$B\times C_l\times H_l/2\times W_l/2$. Thus, the routed $LL$ tensor preserves
the channel count while reducing each spatial dimension by two. A convolutional
block then maps the routed $LL$ tensor to $C_{l+1}$ channels before the next
transition. The low-frequency tensor and the three detail tensors are kept in
the transition cache together with the pooled routing map.

The U-Net backbone uses double-convolution blocks. For nnU-Net, MoWE retains
the configured convolutional blocks,
deep supervision, training procedure, inference, and postprocessing while
replacing the encoder and decoder resolution transitions. In both backbones,
the decoder first projects the deeper representation back to $C_l$ channels
before synthesis. Each expert-specific IDWT
then receives this decoder $LL$ tensor and the cached detail triplet. The
resulting reconstructions are mixed in image space using the nearest-neighbor
upsampled routing map and concatenated with the matching encoder skip tensor.
Fixed DWT/IDWT
filters are registered as non-trainable buffers; only
the router, normalization, adapter, and backbone parameters are optimized.

MoWE evaluates all fixed experts at each transition. The
router selects one expert at each pooled spatial location, but the transform
bank is computed densely so that the hard assignment remains differentiable
during training through the straight-through estimator; Sec.~\ref{sec:rationale}
discusses this design choice and its computational consequences.

\subsection{Context-Aware Spatial Wavelet Routing (Encoder)}\label{sec:encoder}
Given an encoder feature map $X_l \in \mathbb{R}^{C_l \times H_l \times W_l}$ at resolution level $l$, we project it through a bank of $K$ fixed wavelet families (e.g., Haar, db2, db4). Each expert $k \in \{1,\dots,K\}$ computes the Discrete Wavelet Transform (DWT) \cite{alessio2015discrete}, yielding one low-frequency and three high-frequency subbands:
\begin{equation}
  (LL_l^{(k)}, LH_l^{(k)}, HL_l^{(k)}, HH_l^{(k)}) = \mathcal{W}^{(k)}_{\mathrm{DWT}}(X_l).
\end{equation}

Standard mixture-of-experts layers route tokens independently, whereas US textures call for global structural awareness. A $3{\times}3$ convolutional router $\phi_l$ produces per-pixel logits for $K$ experts, which are average-pooled to the subband resolution and biased by a global context vector $\gamma_l$:
\begin{equation}
  \tilde{Z}_l(u) = \mathrm{Pool}(\phi_l(X_l))(u) + \lambda_{\mathrm{ctx}} \gamma_l(\mathrm{GAP}(X_l)),
\end{equation}
where $u$ indexes the spatial location, $\mathrm{Pool}$ denotes $2{\times}2$ average pooling, $\mathrm{GAP}$ is global average pooling, $\gamma_l$ is a lightweight MLP with hidden dimension $C_l/8$, and $\lambda_{\mathrm{ctx}}$ regulates the global context influence. Pooling the logits by the DWT's downsampling factor aligns each assignment with exactly one transform output cell, so routing decisions live at the same granularity as the subbands they select, while the global term lets coarse anatomy bias assignments that local texture alone would leave ambiguous.

To maintain gradient flow while forcing discrete spatial assignments, we use a straight-through Gumbel-Softmax \cite{jang2017gumbel} estimator:
\begin{equation}
  G_l = \mathrm{GumbelSoftmax}(\tilde{Z}_l; \tau, \mathrm{hard}=1),
\end{equation}
where $G_l \in \{0,1\}^{K \times \frac{H_l}{2} \times \frac{W_l}{2}}$ is a one-hot spatial assignment map. The routed low-frequency representation $\hat{LL}_l = \sum_{k=1}^{K} G_l^{(k)} \odot LL_l^{(k)}$ is batch-normalized and forwarded to the next encoder block. Each expert's detail subbands are independently batch-normalized prior to caching. The detail subbands $\mathcal{D}_l^{(k)}=\{LH_l^{(k)},HL_l^{(k)},HH_l^{(k)}\}$ and the spatial assignment map $G_l$ are cached for decoder-side synthesis.

\subsection{Detail-Gated Synthesis and Routing Reuse (Decoder)}\label{sec:decoder}
During reconstruction, the upsampled decoder features~$U_l$ are projected to
the low-frequency dimension, yielding~$LL_l^{\mathrm{dec}}$.
Analysis--synthesis consistency requires that each location be reconstructed with the same wavelet used for decomposition. We therefore reuse the encoder's routing map via nearest-neighbor upsampling,
$\bar{G}_l \!=\! \mathrm{Up}_{\mathrm{nn}}\!(G_l)$,
preserving the hard one-hot structure.

High-frequency subbands $\mathcal{D}_l^{(k)}$ carry lesion boundary detail but also encode US speckle noise. We introduce a Detail Adapter Bank with learned channel-wise gains that rescale these subbands prior to the Inverse DWT (IDWT). For each subband $d \in \mathcal{D}_l^{(k)}$, the adapter computes:
\begin{equation}
\tilde{d} = \sigma(\Theta_{l, d}) \odot d,
\end{equation}
where $\Theta_{l,d} \in \mathbb{R}^{C_l \times 1 \times 1}$ are learnable logits initialized to 2.2, and $\sigma$ is the sigmoid function. The rescaled details are transformed via IDWT:
\begin{equation}
R_l^{(k)} = \mathcal{W}^{(k)}_{\mathrm{IDWT}}\!\left(LL_l^{\mathrm{dec}}, \tilde{\mathcal{D}}_l^{(k)}\right),
\end{equation}
and combined using the reused spatial routing: $\hat{U}_l = \sum_{k=1}^{K} \bar{G}_l^{(k)} \odot R_l^{(k)}$. $\hat{U}_l$ is then fused with skip connections and refined.

\subsection{Optimization and Load Balancing}\label{sec:optimization}
To prevent representational collapse, where the network degenerates to routing all regions to a single wavelet family, we incorporate a Switch-Transformer load balancing regularizer \cite{fedus2022switch}. Let $f_l^{(k)} = \frac{1}{|\Omega|} \sum_{u} G_l^{(k)}(u)$ be the fraction of spatial tokens routed to expert $k$ at level $l$, and $P_l^{(k)} = \frac{1}{|\Omega|} \sum_{u} \mathrm{Softmax}(\tilde{Z}_l(u))^{(k)}$ be the mean routing probability. The total objective is:
\begin{equation}
\mathcal{L} = \tfrac{1}{2}\mathcal{L}_{\mathrm{BCE}} + \mathcal{L}_{\mathrm{Dice}} + \lambda_{\mathrm{bal}} \sum_{l=1}^L \left( K \sum_{k=1}^K f_l^{(k)} P_l^{(k)} \right),
\end{equation}
where $\lambda_{\mathrm{bal}}$ scales the balance penalty. To encourage exploration during early training and sharp expert specialization later, we use a $T_w$-epoch warmup and linearly anneal the Gumbel temperature from $\tau_{\max}$ to $\tau_{\min}$ over the remaining $T-T_w$ epochs: $\tau_t = \tau_{\max} - (\tau_{\max} - \tau_{\min})\,(t-T_w)/(T-T_w)$.

\subsection{Design Rationale}\label{sec:rationale}
MoWE keeps the expert bank fixed so that each expert remains an orthonormal DWT/IDWT pair with known frequency responses and a perfect-reconstruction guarantee~\cite{daubechies1992ten}. A routing decision can therefore be interpreted as selecting a locally appropriate frequency decomposition, while the bank adds no trainable parameters. A learned transform would obscure this interpretation and would also need to satisfy reconstruction constraints throughout optimization.

This interpretation depends on hard assignments. Softly mixing subbands from different orthonormal bases produces a representation that lies in no single basis and makes a reconstruction difficult to attribute to one expert. We therefore use hard assignments with a straight-through estimator, which keeps the routing discrete while preserving gradient-based training.

The decoder must use the same basis that was selected during analysis. Reconstructing a region with a different basis breaks the analysis/synthesis pairing and introduces a systematic approximation error at every transition. MoWE reuses the encoder's assignment map during decoding. Nearest-neighbor upsampling preserves its one-hot structure, so synthesis uses the exact basis selected during analysis without adding decoder-side routing cost.

These choices require dense expert evaluation. Every transition computes all $K$ transforms and caches all detail triplets, so transform compute and activation memory grow linearly in $K$ relative to a single-family wavelet U-Net. The fixed transforms replace the baseline downsampling and upsampling operators: max-pooling or strided convolution in the encoder and transposed convolution in the decoder. Trainable parameters therefore decrease by 6.4\% in the U-Net configuration (Sec.~\ref{sec:efficiency}); the measured latency and memory overhead are reported there.

\section{Experimental Setup}\label{sec:experiments}

\textbf{Datasets and Preprocessing.} We evaluate on BUSI (breast) \cite{al2020dataset}, UDIAT (breast) \cite{yap2017automated}, BUS-UCLM (breast) \cite{vallez2025bus}, TN3K (thyroid) \cite{gong2023thyroid}, and CAMUS (cardiac) \cite{leclerc2019deep}. Table~\ref{tab:datasets} summarizes the resulting corpus. All images are resized to $256\times256$ pixels. For BUSI, UDIAT, and BUS-UCLM, images are read as single-channel intensity maps and divided by 255; masks are resized with nearest-neighbor interpolation. BUSI excludes the normal class and filters samples with missing or empty lesion masks. TN3K uses the official trainval folds and held-out test set, with masks binarized at intensity threshold 127. CAMUS retains complete patients with 2CH/4CH end-diastolic and end-systolic frames, min--max normalizes each image independently, and represents background, left ventricle, myocardium, and left atrium as four mutually exclusive classes.

\begin{table}[t]
\centering
  \caption{Dataset statistics after preprocessing. Splits list train/val/test. CAMUS counts patients; each contributes four annotated frames. TN3K lists total trainval/test sizes; per-fold splits follow the official five folds. UDIAT splits vary by up to one image across folds.}
\label{tab:datasets}
\small
\setlength{\tabcolsep}{4pt}
\begin{tabular}{llll}
  \toprule
  Dataset & Anatomy & Size & Split protocol \\
  \midrule
  BUSI & breast & 647 & 388/129/130 holdout \\
  UDIAT & breast & 163 & 5-fold CV (111/20/32) \\
  BUS-UCLM & breast & 683 & 417/126/140 patient \\
  TN3K & thyroid & 3493 & 2879/614, official folds \\
  CAMUS & cardiac & 500 pat. & 300/100/100 patient \\
  \bottomrule
\end{tabular}
\end{table}

BUSI uses a 60/20/20 image-level holdout; UDIAT uses five folds with a validation fraction of 0.15 within each training fold; BUS-UCLM uses patient-level grouping; TN3K uses its five predefined folds; and CAMUS uses a 60/20/20 patient-level split. BUSI, BUS-UCLM, and CAMUS are evaluated over five random seeds, while UDIAT and TN3K are evaluated over five folds.

Training uses batch size 28 with random 90-degree rotations and flips. Validation and test samples are evaluated without augmentation.

\textbf{Implementation Details.} Models are implemented in PyTorch, and computational profiling is performed on an NVIDIA RTX 3090 GPU (24 GB). The fixed expert bank is $K=3$ with Haar, db2, and db4 filters. The optimizer is Adam with learning rate $10^{-4}$ and weight decay $10^{-4}$; the cosine schedule decays to $10^{-5}$ over 200 epochs. Training uses automatic mixed precision and lasts 200 epochs. The binary objective is $\tfrac{1}{2}\mathcal{L}_{\mathrm{BCE}} + \mathcal{L}_{\mathrm{Dice}}$; CAMUS uses multiclass cross-entropy plus foreground Dice. The router uses a five-epoch uniform warmup, context reduction factor 8, context scale 0.5, balance weight 0.005, and linear temperature annealing from 1.0 to 0.5. Low-frequency and detail batch normalization are enabled, and detail-adapter logits are initialized to 2.2.

\textbf{Model selection and inference.} Baseline and MoWE runs with matching seeds reused the same data split. We fixed the seed for weight initialization, data shuffling, and augmentation. During validation and testing, we assigned each location to the expert with the largest router logit and did not sample Gumbel noise. The fixed DWTs use circular boundary extension; the matching inverse transforms fold the periodic samples back into the reconstruction.

\textbf{nnU-Net comparison.} We additionally evaluate nnU-Net~\cite{isensee2021nnu}, a strong and widely used medical image segmentation benchmark, on BUSI. The baseline and nnU-Net + MoWE use the same nnU-Net configuration, data split, training procedure, inference, and postprocessing; MoWE changes the resolution transitions.

\textbf{Metric computation.} Binary predictions are obtained by applying a sigmoid and thresholding at 0.5. Dice and IoU are computed per case and then averaged; reported variation is calculated across training seeds or predefined folds. HD95 and average symmetric surface distance (ASSD) are computed with MedPy, with the image diagonal assigned when exactly one of the prediction and target masks is empty. CAMUS predictions use softmax followed by argmax, and foreground metrics are macro-averaged over the three anatomical classes.

\begin{table}[t]
\centering
\caption{Quantitative U-Net results on ultrasound lesion datasets. Values are mean$\pm$std over five seeds for BUSI and BUS-UCLM and five folds for UDIAT and TN3K. DSC in \%, HD95 in px; the better result is \textbf{bolded}.}
\label{tab:res_lesions}
\setlength{\tabcolsep}{3pt}
\small
\begin{tabular}{lcccc}
  \toprule
  & \multicolumn{2}{c}{DSC $\uparrow$} & \multicolumn{2}{c}{HD95 $\downarrow$} \\
  \cmidrule(lr){2-3} \cmidrule(lr){4-5}
  Dataset & U-Net & +MoWE & U-Net & +MoWE \\
  \midrule
  BUSI     & \vstd{72.59}{1.51}          & \vstd{\textbf{74.55}}{1.66} & \vstd{47.01}{3.65}          & \vstd{\textbf{43.49}}{3.01} \\
  UDIAT    & \vstd{\textbf{74.65}}{5.44} & \vstd{73.29}{1.78}          & \vstd{\textbf{29.72}}{7.00} & \vstd{34.64}{4.80}          \\
  BUS-UCLM & \vstd{73.48}{2.46}          & \vstd{\textbf{74.48}}{2.24} & \vstd{43.53}{6.15}          & \vstd{\textbf{39.63}}{5.80} \\
  TN3K     & \vstd{79.73}{0.93}          & \vstd{\textbf{81.19}}{0.53} & \vstd{32.92}{2.20}          & \vstd{\textbf{29.90}}{1.56} \\
  \bottomrule
\end{tabular}
\end{table}
\section{Results and Discussion}
\subsection{U-Net Results} We first evaluate MoWE as a replacement for the standard resolution transitions in U-Net. Performance is quantified using the Dice Similarity Coefficient (DSC) for regional overlap and the 95th-percentile Hausdorff Distance (HD95) for boundary precision. The results are reported in Table~\ref{tab:res_lesions}.

On BUSI, MoWE raises the mean DSC from 72.59\% to 74.55\% and reduces HD95 from 47.01 to 43.49\,px. On BUS-UCLM, it raises DSC from 73.48\% to 74.48\% and reduces HD95 from 43.53 to 39.63\,px. On TN3K, DSC increases from 79.73\% to 81.19\% and HD95 decreases from 32.92 to 29.90\,px. UDIAT shows the opposite trend: DSC decreases from 74.65\% to 73.29\%, while HD95 increases from 29.72 to 34.64\,px.

UDIAT is the only dataset where MoWE lowers the mean. MoWE reduces the cross-fold standard deviation of DSC from 5.44 to 1.78 points while shifting the mean by $-1.36$ points, so the routed model trades a small average loss for markedly more consistent predictions. UDIAT is also the smallest benchmark in our study: 163 images from a single center, with roughly 111 training images per fold (Table~\ref{tab:datasets}). The result is consistent with the possibility that the additional routing components are harder to estimate in this setting, although the current experiments do not isolate the effects of dataset size or acquisition heterogeneity. The worsened HD95 alongside the mild DSC change suggests that the regression concerns boundary localization rather than gross region failure.

\subsection{Comparison with nnU-Net} nnU-Net is a strong benchmark across diverse medical image segmentation tasks~\cite{isensee2021nnu,isensee2024nnunet_revisited}. We therefore compare nnU-Net with nnU-Net + MoWE on BUSI (Table~\ref{tab:res_nnunet}).

\begin{table}[t]
  \centering
  \small
  \setlength{\tabcolsep}{5pt}
  \caption{Comparison with nnU-Net on BUSI. DSC and IoU are reported in \%; ASSD and HD95 are reported in px.}
  \label{tab:res_nnunet}
  \begin{tabular}{lcccc}
    \toprule
    Method & DSC $\uparrow$ & IoU $\uparrow$ & ASSD $\downarrow$ & HD95 $\downarrow$ \\
    \midrule
    nnU-Net & 80.18 & 71.89 & 20.07 & \textbf{52.34} \\
    nnU-Net + MoWE & \textbf{81.97} & \textbf{73.58} & \textbf{17.25} & 54.02 \\
    \bottomrule
  \end{tabular}
\end{table}

MoWE improves DSC by 1.79 percentage points and IoU by 1.69 points, while reducing ASSD from 20.07 to 17.25\,px (a 14.1\% reduction). HD95 changes from 52.34 to 54.02\,px; the gains lie in overlap and average surface agreement.

ASSD measures the mean symmetric surface distance within each case, whereas HD95 measures the 95th percentile of the surface distances; both are then averaged across cases. Their divergence suggests that average surface agreement improves while tail errors remain, although the aggregate values do not establish the per-case failure pattern. The routing distribution also changes with the backbone: with nnU-Net, the router assigns 46.8\% of pooled locations to db2 and only 19.5\% to db4, a less balanced distribution than under U-Net (Sec.~\ref{sec:efficiency}).

\subsection{Cross-Domain Evaluation} We also evaluate on CAMUS echocardiography (Table~\ref{tab:res_camus}). On U-Net, DSC changes from 89.14\% to 89.10\%, while HD95 decreases from 8.38 to 7.99\,px.

Echocardiography probes a different operating regime than lesion segmentation: chamber boundaries are large and supported by strong shape priors, the baseline already exceeds 89\% DSC, and speckle interacts with smooth chamber walls rather than ill-defined lesion margins. In this regime MoWE leaves overlap unchanged ($-0.04$ points) while improving boundary precision (HD95 $-0.39$\,px, a 4.7\% reduction) and reducing variance (DSC std 0.66 to 0.24, HD95 std 0.68 to 0.26). Outside its target domain, MoWE leaves an already-strong baseline intact and concentrates its benefit on boundary precision.

\begin{figure*}[t]
\centering
\setlength{\tabcolsep}{1.2pt}
\renewcommand{\arraystretch}{0.8}
\begin{tabular}{@{}ccccc@{}}
  \scriptsize Image & \scriptsize GT & \scriptsize Base & \scriptsize MoWE & \scriptsize Route \\
  \multicolumn{5}{c}{\scriptsize\textbf{BUSI}} \\
  \includegraphics[width=0.185\textwidth]{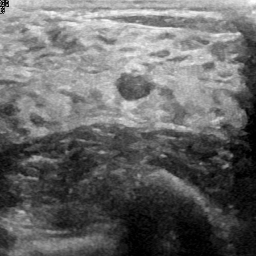} &
  \includegraphics[width=0.185\textwidth]{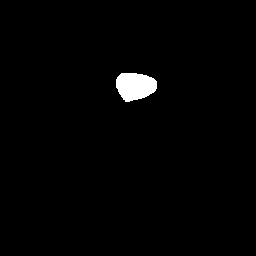} &
  \includegraphics[width=0.185\textwidth]{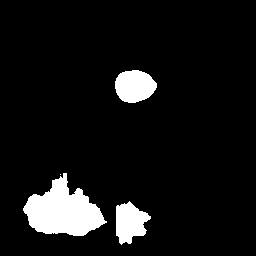} &
  \includegraphics[width=0.185\textwidth]{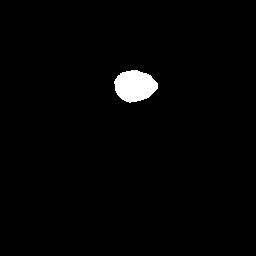} &
  \includegraphics[width=0.185\textwidth]{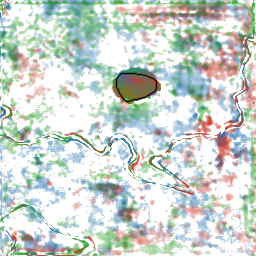} \\
  \multicolumn{5}{c}{\scriptsize\textbf{UDIAT}} \\
  \includegraphics[width=0.185\textwidth]{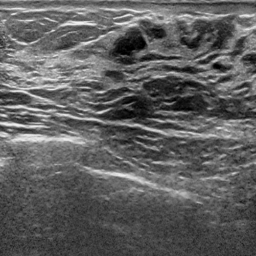} &
  \includegraphics[width=0.185\textwidth]{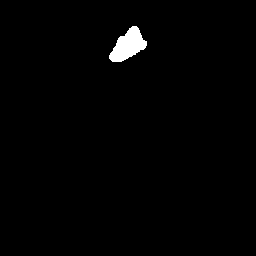} &
  \includegraphics[width=0.185\textwidth]{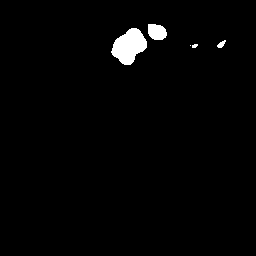} &
  \includegraphics[width=0.185\textwidth]{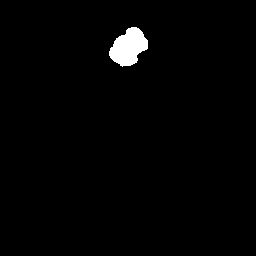} &
  \includegraphics[width=0.185\textwidth]{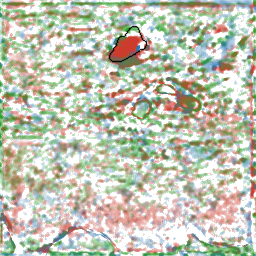} \\
  \multicolumn{5}{c}{\scriptsize\textbf{BUS-UCLM}} \\
  \includegraphics[width=0.185\textwidth]{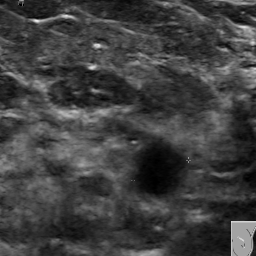} &
  \includegraphics[width=0.185\textwidth]{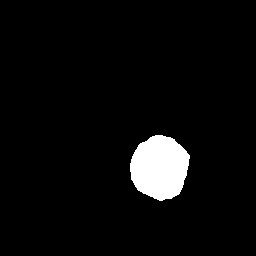} &
  \includegraphics[width=0.185\textwidth]{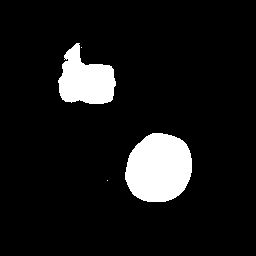} &
  \includegraphics[width=0.185\textwidth]{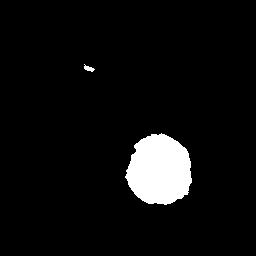} &
  \includegraphics[width=0.185\textwidth]{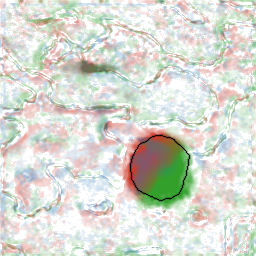} \\
  \multicolumn{5}{c}{\scriptsize\textbf{TN3K}} \\
  \includegraphics[width=0.185\textwidth]{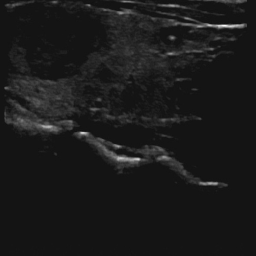} &
  \includegraphics[width=0.185\textwidth]{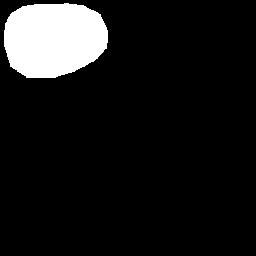} &
  \includegraphics[width=0.185\textwidth]{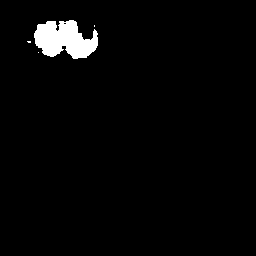} &
  \includegraphics[width=0.185\textwidth]{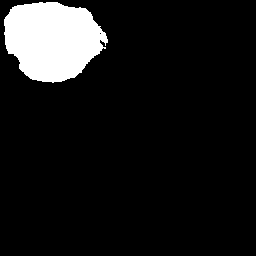} &
  \includegraphics[width=0.185\textwidth]{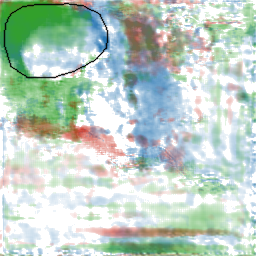} \\
\end{tabular}
\caption{Qualitative U-Net comparison. Standard spatial pooling can miss boundaries in regions affected by acoustic shadowing. MoWE produces tighter boundaries in these examples. The Route column shows the expert-contribution map, with Haar, db2, and db4 encoded as \textbf{red}, \textbf{blue}, and \textbf{green}, respectively.}
\label{fig:qualitative_results}
\end{figure*}

\subsection{Qualitative Analysis} Figure~\ref{fig:qualitative_results} shows representative U-Net predictions and expert-contribution maps. Standard spatial pooling can cause the predicted mask to extend into surrounding tissue in regions affected by acoustic shadowing. MoWE produces tighter boundaries in these examples. For each expert $k$, the contribution map is computed from the signed difference between the full-model logits and the logits obtained when expert $k$ is excluded. Positive contributions are normalized across experts to obtain a per-pixel share, visualizing the spatial influence of each expert on the prediction.

\begin{table}[t]
  \centering
  \setlength{\tabcolsep}{3pt}
  \small
  \caption{Wavelet U-Net controls on BUSI. Each row applies a single fixed wavelet family at every resolution transition; all other components match the MoWE configuration. DSC and IoU are reported in \%; HD95 and ASSD are reported in px.}
  \label{tab:fixed_wavelets}
  \begin{tabular}{lcccc}
    \toprule
    Family & DSC $\uparrow$ & IoU $\uparrow$ & HD95 $\downarrow$ & ASSD $\downarrow$ \\
    \midrule
    Haar & \vstd{73.47}{2.19} & \vstd{64.79}{1.95} & \vstd{\textbf{41.66}}{2.19} & \vstd{\textbf{18.76}}{2.60} \\
    db2  & \vstd{\textbf{73.76}}{2.39} & \vstd{\textbf{65.14}}{2.13} & \vstd{41.95}{2.77} & \vstd{20.12}{3.74} \\
    db4  & \vstd{73.69}{2.86} & \vstd{64.97}{2.41} & \vstd{44.01}{3.57} & \vstd{20.77}{4.40} \\
    \bottomrule
  \end{tabular}
\end{table}

\begin{table}[t]
  \centering
  \small
  \setlength{\tabcolsep}{4pt}
  \caption{Component ablation on BUSI using the U-Net backbone.}
  \label{tab:ablation}
    \begin{tabular}{l ccc cc}
      \toprule
      \textbf{Method} & \textbf{$K$} & \textbf{Ctx.} & \textbf{Det.} & \textbf{DSC} $\uparrow$     & \textbf{HD95} $\downarrow$  \\
      \midrule
      U-Net           & -            & -             & -             & \vstd{72.59}{1.51}          & \vstd{47.01}{3.65}          \\
      U-Net (MoWE)    & 1            & -             & \cmark        & \vstd{73.47}{2.19}          & \vstd{41.66}{2.19}          \\
      \midrule
      U-Net (MoWE)    & 3            & \xmark        & \cmark        & \vstd{73.61}{2.53}          & \vstd{45.03}{6.52}          \\
      U-Net (MoWE)    & 3            & \cmark        & \xmark        & \vstd{73.45}{2.78}          & \vstd{41.31}{2.80}          \\
      U-Net (MoWE)    & 3            & \xmark        & \xmark        & \vstd{73.19}{1.67}          & \vstd{\textbf{41.00}}{2.94} \\
      \midrule
      U-Net (MoWE)    & 3            & \cmark        & \cmark        & \vstd{\textbf{74.55}}{1.66} & \vstd{43.49}{3.01}          \\
      \bottomrule
    \end{tabular}
\end{table}

\begin{table}[t]
  \centering
  \small
  \setlength{\tabcolsep}{6pt}
  \caption{Cross-domain results on CAMUS echocardiography.}
  \label{tab:res_camus}
    \begin{tabular}{l cc}
      \toprule
      & DSC $\uparrow$              & HD95 $\downarrow$          \\
      \midrule
      U-Net          & \vstd{\textbf{89.14}}{0.66} & \vstd{8.38}{0.68}          \\
      U-Net (MoWE)   & \vstd{89.10}{0.24}          & \vstd{\textbf{7.99}}{0.26} \\
      \bottomrule
    \end{tabular}
\end{table}

\subsection{Comparison with Fixed Wavelet U-Nets}\label{sec:fixed_wavelets}
Table~\ref{tab:fixed_wavelets} isolates the contribution of spatial routing from the contribution of wavelet downsampling itself. Each control is a Wavelet U-Net that applies a single wavelet family at every resolution transition; the architecture, training protocol, and detail adapter are otherwise identical to MoWE, and the routed model differs only in selecting among the three families per location. The $K\!=\!1$ row of Table~\ref{tab:ablation} (73.47\% DSC, 41.66\,px HD95) coincides with the Haar control.

The fixed controls show an overlap--boundary trade-off. db2 attains the best fixed-family overlap (DSC 73.76\%, IoU 65.14\%), whereas Haar attains the best boundary scores (HD95 41.66\,px, ASSD 18.76\,px); DSC varies by only half a point across families while HD95 varies by 2.35\,px. This split is consistent with the filters' support length and vanishing moments~\cite{daubechies1992ten}: Haar's two-tap support can favor boundary localization, while db2's longer, smoother filter can represent gradually varying tissue more compactly. No single fixed family is simultaneously best on both metric families, indicating the utility of adaptive spatial routing.

The routed model shifts this trade-off toward overlap: it reaches 74.55\% DSC, 0.79 points above the strongest fixed control, while its HD95 (43.49\,px) lies between the fixed extremes. Fixed Haar still surpasses the routed model on HD95: when boundary quality alone matters and the dataset is known in advance, a single compact basis is a strong choice, whereas routing attains the best overlap without requiring that choice a priori. The mean BUSI routing shares (Haar 31.6\%, db2 35.4\%, db4 33.1\%) show that the router draws on all three families rather than collapsing onto one (Sec.~\ref{sec:efficiency}).

\subsection{Component Ablation}
Table~\ref{tab:ablation} then reports an additive component study on BUSI using the U-Net backbone. A single-expert Haar variant improves DSC from 72.59\% to 73.47\% over the U-Net baseline. The full $K\!=\!3$ model reaches 74.55\% DSC and 43.49\,px HD95, separating the use of a fixed wavelet transition from routing among wavelet families.

We then ablate each auxiliary component from the full $K\!=\!3$ model. Removing the context bias ($\lambda_{\mathrm{ctx}}\!=\!0$; DSC: 73.61\%, HD95: 45.03\,px) leaves routing based on the local $3{\times}3$ logits. Without the Detail Adapter (DSC: 73.45\%, HD95: 41.31\,px), the high-frequency subbands are passed to reconstruction without learned rescaling, improving HD95 at the expense of DSC. Removing both components produces the weakest $K\!=\!3$ overlap (DSC 73.19\%) yet the best boundary score of the ablation (HD95 41.00\,px). This pattern mirrors the fixed-family trade-off of Sec.~\ref{sec:fixed_wavelets}: boundary sharpness and region overlap pull in different directions, while the full combination attains the highest DSC among the ablations.

\subsection{Computational Cost and Routing Diagnostics}\label{sec:efficiency}
MoWE evaluates every expert transform densely before applying the hard routing
mask. GFLOPs include convolutions within the DWT and IDWT; latency is measured
at $256\times256$ with batch size one. The resulting trade-off is summarized in
Table~\ref{tab:efficiency}. On U-Net, MoWE reduces trainable parameters by
6.4\% because the fixed DWT/IDWT filters replace max-pooling and
transposed-convolution upsampling, while forward latency increases by 3.21$\times$
and training-step latency by 2.36$\times$ in this measurement. GFLOPs remain nearly unchanged (96.18 to 96.45)
because the wavelet transforms themselves are cheap; the latency gap reflects
the memory traffic of caching all expert subbands rather than arithmetic cost.

\begin{table}[t]
  \centering
  \setlength{\tabcolsep}{4pt}
  \small
  \caption{Single-image computational profile at $256\times256$ on an RTX 3090 GPU (50 timed repetitions). Parameters are in millions; Fwd and Step are latency in ms; Mem is peak training memory in MB.}
  \label{tab:efficiency}
  \begin{tabular}{lrrrrr}
    \toprule
    Model & Params & GFLOPs & Fwd & Step & Mem \\
    \midrule
    U-Net & 31.04 & 96.18 & 5.37 & 19.35 & 728.0 \\
    U-Net (MoWE) & 29.04 & 96.45 & 17.24 & 45.62 & 1124.2 \\
     \bottomrule
  \end{tabular}
\end{table}

Routing statistics reveal how expert selection depends on the backbone and on
encoder depth. With U-Net on BUSI, the mean shares are balanced
(Haar 31.6\%, db2 35.4\%, db4 33.1\%), and the load-balancing term keeps every
expert well above the collapse regime. Assignments nevertheless shift
systematically with depth: the db2 share decreases monotonically from 36.4\% at
the shallowest transition to 28.4\% at the deepest, while the Haar share rises
from 31.5\% to 36.8\% and db4 peaks mid-encoder (37.1\% at the second-deepest
transition). One possible interpretation is that shallower texture-heavy features
favor smoother bases, whereas deeper features favor the compact Haar basis;
the usage counts alone, however, do not identify which feature properties drive
these assignments. Under nnU-Net on the same dataset, the distribution changes
markedly (db2 46.8\%, Haar 33.6\%, db4 19.5\%), showing that the selected mixture
also depends on the backbone under the same task.

\subsection{Limitations}
The current design has practical and empirical limitations. Dense evaluation of the expert bank makes the method more expensive: forward latency increases by 3.21$\times$, training-step latency by 2.36$\times$, and peak training memory by 1.54$\times$ over U-Net despite the lower parameter count (Table~\ref{tab:efficiency}). Evaluating only the selected expert would reduce this overhead, but would remove the dense gradients that make the hard assignment trainable with a straight-through estimator. The gains also vary across datasets: on the small, single-center UDIAT corpus, the routed model has lower mean accuracy, while with nnU-Net it does not improve the tail boundary metric (HD95). Finally, the evaluation covers only 2D ultrasound and echocardiography, uses $K\!=\!3$ Daubechies-family experts, and is limited to U-Net and nnU-Net. Additional U-Net variants, larger expert banks, learnable bases, and 3D volumetric data remain to be evaluated.

\section{Conclusion}
Ultrasound segmentation presents a difficult trade-off at every resolution transition: preserving weak boundary evidence can also preserve speckle, while suppressing noise can erase lesion structure. Our results suggest that this trade-off should not be resolved by choosing one pooling operator or wavelet basis for the entire image. MoWE instead adapts the frequency decomposition to spatial content while maintaining consistent analysis and reconstruction. On BUSI, the fixed-wavelet controls exposed a clear split between bases that favor overlap and those that favor boundary localization; spatial routing achieved the best overlap among them. This advantage extended beyond the controlled comparison: MoWE improved both Dice and HD95 on three of four lesion datasets with U-Net, and improved Dice and ASSD when integrated with nnU-Net on BUSI, while using 6.4\% fewer trainable parameters. The evidence is not uniformly positive: performance declined on the small, single-center UDIAT dataset, nnU-Net HD95 did not improve, and dense expert evaluation substantially increased latency. The central finding is therefore not that one wavelet is a better pooling operator, but that the appropriate resolution transition can vary across spatial regions and network contexts. Efficient sparse routing, determination of the best wavelet family and number of wavelets, evaluation on various U-Net architectures and larger datasets should be the next step towards assessing whether this principle generalizes beyond our present setting.

\bibliographystyle{ieeenat_fullname}
\bibliography{main}

\end{document}